\documentclass{article}
\usepackage{ap_2017}
\usepackage{mylingmacros} 

\usepackage[utf8]{inputenc} 
\usepackage[T1]{fontenc}    

\let\quoteOLD\quote
\def\quote{\quoteOLD\small}

\usepackage{amsmath}
\usepackage{graphicx}
\usepackage{subcaption}
\usepackage[usenames, dvipsnames]{color}

\usepackage{hyperref}       
\usepackage{url}            
\usepackage{booktabs}       
\usepackage{amsfonts}       
\usepackage{nicefrac}       
\usepackage{microtype}      
\usepackage{natbib}
\usepackage{soul}

\title{Edina: Building an Open Domain Socialbot with Self-dialogues}

\author{
    Ben Krause \ \ \ Marco Damonte$^\star$ \ \ \ Mihai Dobre$^\star$ \ \ \ Daniel Duma$^\star$ \ \ \ Joachim Fainberg$^\star$ \\ \textbf{Federico Fancellu}$^{\star\dagger}$ \ \ \ \textbf{Emmanuel Kahembwe}$^\star$ \ \ \ \textbf{Jianpeng Cheng} \ \ \ \textbf{Bonnie Webber}$^\ddagger$\\
	School of Informatics\\
	University of Edinburgh\\
	Edinburgh, UK\\
	\texttt{ben.krause@ed.ac.uk, f.fancellu@sms.ed.ac.uk, bonnie@inf.ed.ac.uk}\\ 
}

\begin{document}
\maketitle
\begin{abstract}
We present \textit{Edina}, the University of Edinburgh's social bot for the Amazon Alexa Prize competition. \textit{Edina} is a conversational agent whose responses utilize data harvested from Amazon Mechanical Turk (AMT) through an innovative new technique we call \textit{self-dialogues}. These are conversations in which a single AMT Worker plays both participants in a dialogue. Such dialogues are surprisingly natural, efficient to collect and reflective of relevant and/or trending topics. These self-dialogues provide training data for a generative neural network as well as a basis for \textit{soft rules} used by a matching score component. Each match of a soft rule against a user utterance is associated with a confidence score which we show is strongly indicative of reply quality, allowing this component to self-censor and be effectively integrated with other components. Edina's full architecture features a rule-based system backing off to a matching score, backing off to a generative neural network.  Our hybrid data-driven methodology thus addresses both coverage limitations of a strictly rule-based approach and the lack of guarantees of a strictly machine-learning approach. 
\let\thefootnote\relax\footnotetext{
* equal contribution; $\dagger$ team leader; $\ddagger$ faculty advisor}
\end{abstract}

\section{Introduction}
Building an open-domain conversational AI for commercial use poses two main challenges. First is broad-coverage: Modeling natural conversation in an unrestricted number of topics is still an open problem as shown by the current concentration of research on dialogue in restricted domains (e.g. \cite{bowden2017data}). 
Second is the scarcity of clean, unbiased and comprehensive datasets of open-ended conversation, which makes it difficult to develop conversational dialogue systems and limits the viability of using purely data-driven methods (e.g. neural networks).

Our approach aims to integrate rule-based with machine-learned behavior, with both grounded in \textbf{data} that is \textit{intelligible, reflective of what people want to talk about, and gathered automatically}. This means acknowledging that people talk differently about different things at different times. The \textbf{domain} has to be \textit{potentially} unrestricted and the \textbf{model} should be powerful enough to ensure that the most likely system response is in line with both the user's most recent utterance and, ideally, the flow of the overall conversation. 

We present \textit{Edina}, a conversational AI agent that exploits a corpus of conversations harvested from Amazon Mechanical Turk (\S2). Our innovation is to collect and use data in the form of \textbf{self-dialogues}, in which Workers engage in conversation with themselves on a specific topic, enabling us to gather first-hand what people discuss and how they discuss it. By running tasks periodically, we are able to identify the trending entities people talk about at a given point in time. Although our system architecture (\S3) also includes a rule-based component (\S3.1), our growing corpus of self-dialogues is designed to enable us to avoid having to hand-script potential conversational situations. We achieve this by an intuitive IDF (inverse document frequency)-based matching score (\S 3.3) that takes a user utterance and returns the most likely response, based on either conversational partner in a self-dialogue. The matching score is also able to self-censor when it is not confident that it has an intelligible reply, making it easy to integrate with other components. Although \textit{Edina} currently converses on only three main topics (movies, music and sports), the approach is \textbf{easily extensible} to any topic. Our approach is also \textbf{cost-effective}: Once an initial quantity of data is collected around a topic, only small updates are needed in order to stay current on trending topics. After describing our methods in detail, we perform qualitative and quantitative evaluations of our system (\S \ref{sec:eval}), and describe the significance of our approach and results in the discussion (\S \ref{discussion}).

\section{Data collection}
\begin{figure}[h]
  \centering
  \fbox{\includegraphics[width=0.8\columnwidth]{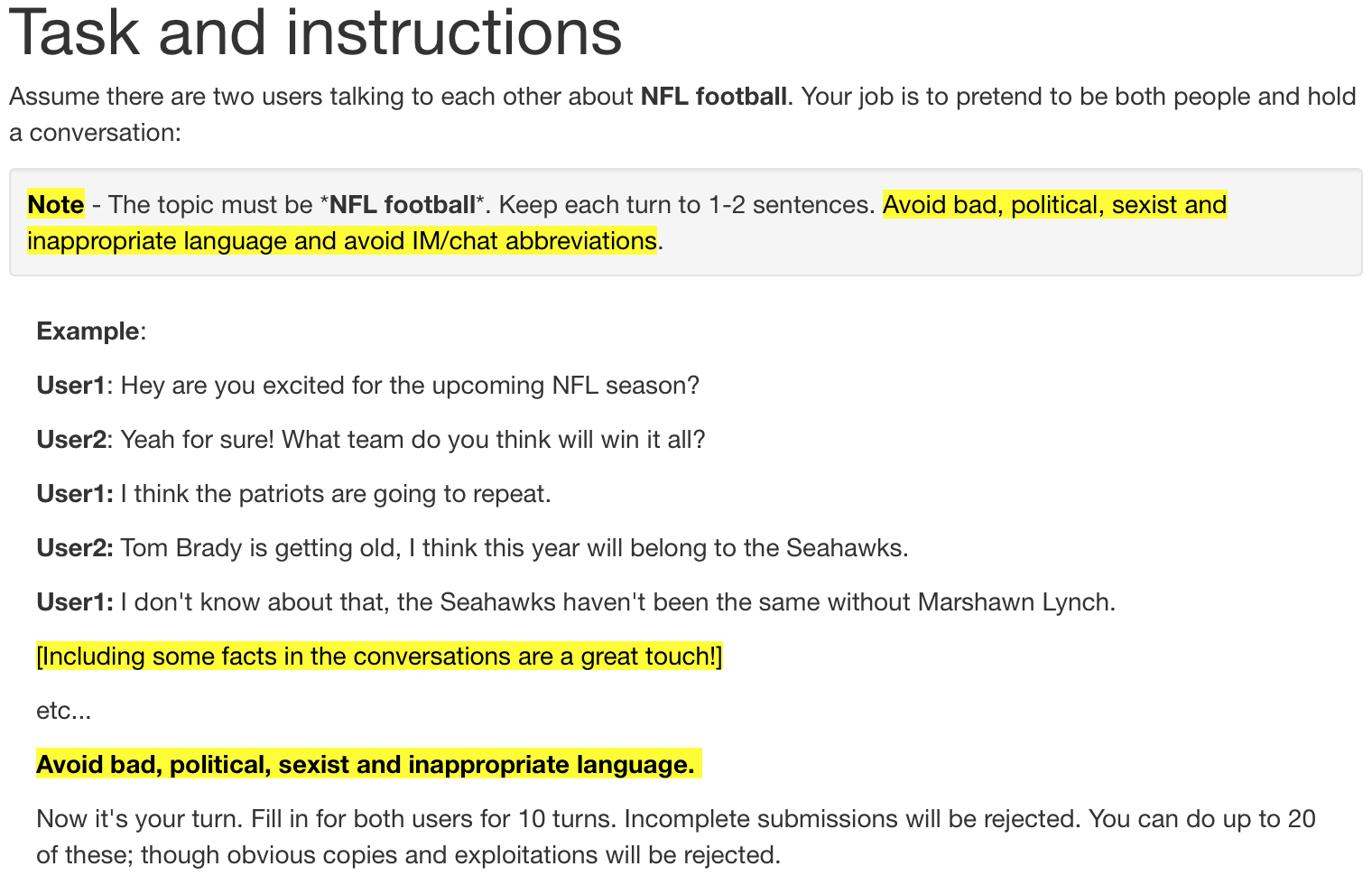}}
  \caption{AMT interface for the NFL Football task.}
  \label{fig:mturk_interface_self}
\end{figure}

Our focus on data collection stems from the scarcity of publicly available corpora for training dialogue systems. During the first stage of the competition, we surveyed the corpora mentioned in \cite{DBLP:journals/corr/SerbanLCP15} and found that suitable publicly available corpora were typically too small, artificial or difficult to obtain. We instead chose to gather our own data with the primary criteria that the conversations i) represent casual human dialogue and ii) have identifiable topics. We turned to Amazon Mechanical Turk (AMT), where we initially developed a chat interface for two AMT Workers to chat about a specific topic. The interface was based on VisDial \citep{DBLP:journals/corr/DasKGSYMPB16}, an interface that prompts the Workers to converse about a particular image. However, in demanding that two workers be connected at the same time, this setup slowed down the collection task on an otherwise fast-moving platform. This led us to develop an innovative new task on AMT, \textbf{self-dialogue}, in which a single worker is asked to play both parts in a conversation about a given topic. A sample self-dialogue is given in Figure \ref{fig:mturk_interface_self},
alongside the instructions provided to the AMT Workers. A key requirement is that the conversations appear natural. For this reason, we avoided lengthy instruction, instead allowing Workers to interpret the task as they wish within the given topic.

Overall we observed the following data-related benefits of self-dialogues, as compared with the two-person conversations that are more standardly collected:
\begin{enumerate}
\item \textbf{Collection speed and efficiency of set up}: Self-dialogues do not require waiting for two workers to connect at the same time, speeding up the collection process and easing the development of the back-end.
\item \textbf{Data quality}: Both parties of a self-dialogue are equally expert in the same topics, making it much easier for self-dialogues to go into detail about particular topics. We also found that two-person dialogues often showed misunderstandings between speakers, and clarifications that play out over several turns, requiring undesirable complexity of the conversation agent.
\item \textbf{Naturalness}: It was striking the extent to which people were able to have interesting and engaging conversations with themselves, adopting different perspectives and even different stances with respect to a topic, depending on which participant they were pretending to be at each point in the conversation. While one might have expected the conversations to simply resemble question-answer pairs, participants instead often contributed a pair of communicative actions --- often with the first part addressing the previous conversation and the other, advancing it further. Such contributions are particularly valuable as data because they have built-in hook to further engagement.
\item \textbf{Cost effectiveness}: Related to (2), we were able to halve the cost of collecting conversation since we only had to pay one worker instead of two.
\end{enumerate}

Submissions were of surprisingly high quality. We implemented a pipeline to review and identify Workers abusing the system or submitting duplicates. However, only eight of 2,717 were banned, and only 145 conversations  ($\approx 0.6\%$) were rejected. Initial experiments with requirements on Workers and the pay per conversation converged on the following set:
\begin{itemize}
\item location: the United States or United Kingdom;
\item HIT approval rate: greater than 95\%;
\item number of HITs approved: greater than 500;
\item number of conversations per worker per task: maximum 20;
\item pay per 10-turn conversation: US \$0.70-80; 5-turn conversation: US \$0.35-40.
\end{itemize}
Workers produced 10-turn dialogues for the main topics, and 5-turn dialogues for sub-topics (except for "NFL Football").  Empirically, we observed that the 10-turn tasks afforded Workers the time to transition into a topic of interest, which was unnecessary for sub-topics.

To date, we have collected 24,283 self-dialogues through AMT, across four major topics separated into 23 tasks. These currently amount to 3,653,313 words, across 141,945 turns, from 2,717 Workers. On average, each worker has submitted $\sim$9 self-dialogues, so the corpus displays a wide variety of styles, content and ideas across a large population.

The total paid to Workers so far is US \$17,947.54. AMT's high liquidity has enabled us to gather data fast when required. In a period of approximately 20 days (from 20/06 to 10/07) we were able to gather about 20,000 conversations, with a peak of 2,307 conversations collected in a single day. The following is an excerpt from a self-dialogue in the Movies category:
\begin{quote}
What is your absolute favorite movie?\\
I think Beauty and the Beast is my favorite.\\
The new one?\\
No, the cartoon.  Something about it just feels magical.\\
It is my favorite Disney movie.\\
What's your favorite movie in general?\\
I think my favorite is The Sound of Music.\\
Really?  Other than cartoons and stuff I can never get into musicals.\\
I love musicals.  I really liked Phantom of the Opera.
\end{quote}
The data (or \textit{offline conversations}) are stored by grouping them by their corresponding tasks, with each line in each conversation as a response. In Section~\ref{match-score:sec}, we explain how a context is associated with each response for use with the Matching Score component. 
\begin{table}[t]

  \label{tab:data}
  \centering
  \begin{small}
  \begin{tabular}{lllll}
    \toprule
    Topic/subtopic & \# Conversations & \# Words & \# Turns \\
    \midrule
    Movies & 4,126 & 814,842 & 82,018 \\
    Action & 414 & 37,037 & 4,140 \\
    Comedy & 414 & 36,401 & 4,140 \\
    Fast \& Furious & 343 & 33,964 & 3,430 \\
    Harry Potter & 414 & 44,220 & 4,140 \\
    Disney & 2,331 & 232,573 & 23,287 \\
    Horror & 414 & 428,33 & 4,138 \\
    Thriller & 828 & 77,975 & 8,277\\
    Star Wars & 1,726 & 178,351 & 17,260 \\
    Superhero & 414 & 40,967 & 4,140 \\
    \midrule
    Music & 4,911 & 924,993 & 98,123 \\
    Pop & 684 & 62,383 & 6,840\\
    Rap / Hip-Hop & 684 & 66,376 & 6,840\\
    Rock & 684 & 63,349 & 6,837\\
    The Beatles & 679 &  68,396 & 6,781 \\
    Lady Gaga & 558 & 49,313 & 5,566\\
   	\midrule
    Music and Movies & 216 & 37,303 & 4,320 \\
    \midrule
    NFL Football & 2,801 &  562,801 & 55,939\\
    \bottomrule
  \end{tabular}
  \end{small}
  \smallskip
    \caption{Data collected on AMT and used for Edina. Each line is  an individual task, i.e. \emph{Movies} is not a combination of the related subcategories.
}
\end{table}

\section{System Architecture}
\begin{figure}[h]
  \centering
  \includegraphics[width=\columnwidth]{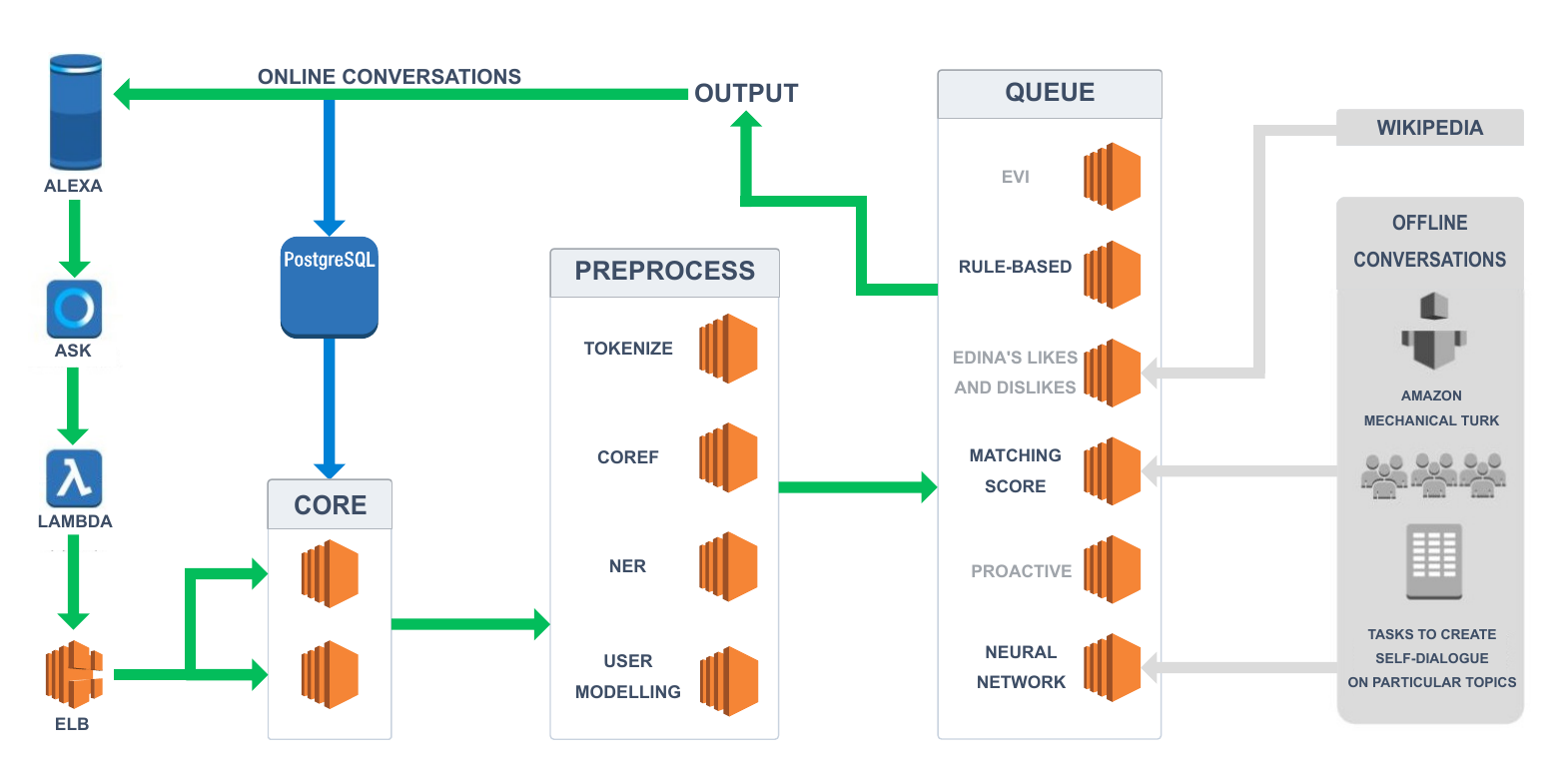}
  \caption{Topological system diagram that shows the order of execution.}
  \label{fig:system_diagram}
\end{figure}

Our system is a hybrid of several rule-based and data-driven components. Following an initial preprocessing step, our system exploits a priority queue of components that vary in their intended coverage and functionality. Its three \textbf{main} components comprise:
\begin{enumerate}
\item[1.] A \textbf{rule-based} component that uses a list of hand written rules and templates. It returns a response only if the user's input is an exact match. While this component has the highest priority, due to obvious coverage limitations, it often returns nothing and defers to the matching score component.
\item[2.] A \textbf{matching score} component that selects responses from a pool of conversational data, based on how close the context of the user conversation is to the context of the response in our data. The matching score also returns a confidence score that is used to better control its interaction with the other components.
\item[3.] A \textbf{generative neural network} that always generates a response and is deployed if the other two components fail. It often gives general and vague on topic responses as compared with the more specific responses of the matching score.
\end{enumerate}
Interleaved between these three main components are three \textit{minor} components:
\begin{enumerate}
\item[4.] \textbf{EVI}, which serves as our primary IR engine, but is only used to reply to user inputs that do not probe the system's identity, personality or opinions. EVI takes priority in responding to WH-questions questions that require an exhaustive knowledge base. As conversations to date have rarely taken this route, EVI has rarely been called.
\item[5.] A \textbf{likes and dislikes} component whose purpose is to answer questions about Edina's opinions on entities and definitions that the rule-based component fails to cover.
\item[6.] A \textbf{proactive component} which asks the user a question or series of questions, in order to steer the conversation back to what the matching score can handle. We limit its use, in order to avoid probing the user too often, resulting in an unpleasant experience.
\end{enumerate}
Although the system is guaranteed to always return something, we retain a set of interesting facts or clarification replies in the unlikely case that all components fail --- say in case of major network errors or if the output from all other components is judged offensive. Output is always filtered for the possibility of profanity. Table~\ref{tab:compfrequency} shows how often each components' output is chosen to be sent back to the Alexa device. Given the order in our priority queue, one can observe the challenges of creating a purely rule-based system that can guarantee full coverage. The rule-based component can only reply to $\approx16\%$ of the user's utterances. Our agent overcomes this limitation by employing data driven methods, i.e. the matching score (\S \ref{match-score:sec}) and generative neural network (\S \ref{sec:nn}) components, that make for the majority of the system's replies ($\approx67\%$). Figure~\ref{fig:system_diagram} presents an overview of the complete system, from input to the Alexa device to the output generated by \textit{Edina} (i.e. green arrow). 


\textit{Edina}'s conversations with Amazon customers as well as other information related to the conversation (e.g. topic, customer preferences, the outputs from each of our components etc.) are stored in a Postgres Amazon RDS instance. 
Throughout the document, we refer to dialogues with Amazon customers as \textit{online conversations} in order to differentiate them from the self-dialogues collected via Amazon Mechanical Turk (\textit{offline conversations}).

\begin{table}[h]
\begin{center} 
\begin{tabular}{  l  l  l  l  l  l  l } \toprule 
RB & MS & NN & EVI & LD & Proactive & Backup \\ 
\midrule 
15.69\% & 46.29\% & 20.55\% & 0.47\% & 0.37\% & 16.56\% & 0.08\% \\  
\bottomrule
\end{tabular} 
\end{center}
\caption{Usage frequency of each component. RB is the rule-based, MS is the matching score, NN is the generative neural network, LD is the likes and dislikes, and Backup is last resort trivia.}
\label{tab:compfrequency}
\end{table}

\subsection{Preprocessing}
We perform a series of preprocessing steps on the raw input received from Lambda. The input is first processed through spaCy's\footnote{\url{https://www.spacy.io}} pipeline mainly for tokenization and Named Entity Recognition (NER). Remaining preprocessing is performed in parallel, comprising another NER, coreference resolution and simple user-modeling. The second NER is performed using DBpedia Spotlight API \citep{daiber2013} which extracts more information on the named entities from Wikipedia. We perform the coreference annotation (COREF) over the previous four turns (i.e. 2 turns each from the user and the bot), using Stanford CoreNLP deterministic implementation \citep{recasens2013}. The generated coreference chain is used to modify the current input message by replacing pronouns with the entities they refer to. Finally, user-modeling is a simple rule-based implementation that catches explicit mentions of the user's likes and dislikes, as well as when the user would like to change the topic. These preferences are matched to a list of topics and subtopics that our system can handle. The processed input and the additional information that result from the preprocessing phase is sent to each of the components that build a reply.

\subsection{Rule-based Component}
The rule-based component deterministically matches a user's input and returns a single output. Rules in the current ruleset address the following points:
\begin{itemize}
\addtolength{\itemsep}{-4pt}
\item \textbf{the agent's identity and preferences}. Rules ensure that identity information remains anonymous for the moment (e.g. name and location). Most preferences are personalized so to make the conversational agent more human-like.
\item \textbf{sensitive topics} such as suicide, cancer or death of a close person, which we carefully redirect to existing helplines when possible. Prompts containing a list of sensitive and/or potentially offensive words are also handled by a polite yet firm response (e.g. `This kind of talk makes me uncomfortable, let's talk about something else.').
\item \textbf{topic shifting}, so as to recognize when the user wants to set a new topic or change the current one, or when the agent should shift away from controversial topics, such as politics, which we do not handle.
\item \textbf{other forms of engagement}. These enable the agent to make jokes, to play a small point-based game, where the user has to complete the lyrics of popular songs, or to invoke a weather API that can return information about the weather, given the user's location.
\end{itemize}
Although the development of the rule-based component was based solely on intuition about what the rules should capture, we have continued to refine our rules and add to our ruleset, based on conversations between Alexa users and \textit{Edina}. Finally, we also integrated the list of frequent and common utterances to all socialbots that have been aggregated and anonymized. The code has been implemented in RiveScript\footnote{\url{https://www.rivescript.com}}.

\subsection{Matching Score component}
\label{match-score:sec}

Given a user utterance $q$, the matching score component is designed to return the most appropriate response $r$ from the bank of self-dialogues (Section \S 2). We treat this bank $B$ as a set of tuples $\{b_1, b_2, ..., b_n\}$ where $b_i$ is a tuple $(r_i, c_i)$ containing a response $r_i$ and a context $c_i$, and $c_i$ is either the immediately preceding response (hence, $c_i$=$r_{i-1}$) or the one at one remove ($c_i$=$r_{i-2}$).
In the context of an online conversation, we define as $q_c$ as the response from \textit{Edina} directly preceding the utterance $q$. This notation is illustrated below:\\

\textbf{Online conversation}\\
previous response from \textit{Edina} ($q_c$): what's your favorite movie \\
user utterance ($q$): sound of music\\
\textbf{Off-line Conversation (from $B$)}\\                                 
$r_{i-2}$: What's your favorite movie in general?    \\  
$r_{i-1}$: I think my favorite is The Sound of Music.   \\    
$r_i$: Really?  Other than cartoons and stuff I can never get into musicals. \\
(where $b_i$ = ($r_i$, $c_i$) and $c_i$ = $r_{i-1}$ $\lor$ $r_{i-2}$)

In order to retrieve the most likely response from $B$, we use a scoring function $\mathcal{S}(q,r_i,c_i)$ that measures the similarity between the user utterance $q$ and the response $r_i$ with a context $c_i$. This similarity measure is based on bag-of-words vectors which up-weight rare words using inverse document frequency (IDF).\footnote{Word frequencies for IDF scores were taken from \url{http://norvig.com/ngrams/count_1w.txt}.} Using bag-of-words permits inverted indexing, where response IDs are stored in a look-up table of words. This removes the need to directly compare $q$ with every $r_i$.

More formally we define our final matching score $\mathcal{S}(q,r_i,c_i)$ as an interpolation of three different subscores plus a normalization term $\eta$:
\begin{equation}
\mathcal{S}(q,r_i,c_i) = \frac{(S^{c}+S^{cr}) (S^{c})^n}{\eta} + \lambda {S}^{2cq}
\end{equation}
where $\lambda$ and $n$ are constant ($\lambda=0.005$ and $n=0.5$).

$S^{c}_i$ is the most important term and measures the similarity between $q$ and $c_i$. Here the context is limited to one preceding response, hence $c_i$=$\{r_{i-1}\}$. Instead of normal IDF scores, we take their cube to give rarer words a higher relative weight. 
Formally, $S^{c}_i$ can be defined as:
\begin{equation}
\mathcal{S}^c = q^3 \cdot c^3_i
\end{equation}
Using this inner product term alone, however, would result in (1) a matching score with a strong preference for responses with high $\lVert c^3_i \rVert_2$, and(2) higher scores for all matches where $\lVert q^3 \rVert_2$ is high. Therefore we introduce a normalization term  
\begin{equation}
\eta =\lVert q^3 \rVert_2 \lVert c^3_i \rVert_2
\end{equation}
Dividing $\mathcal{S}^c_i$ by  $\eta$ would yield a cosine similarity based score.

One shortcoming of $\mathcal{S}^c_i$ is that we might not want to perform exact match on very generic context such as simply "yes". In this case, it might be better for the matching score to defer to another component rather than find a match for "yes" in $B$. It might also be desirable to consider the response $r_i$ when up-weighting overlapping words, because responses that contain words from the query are often more engaging and sometimes more relevant.

$\mathcal{S}^{cr}_i$ addresses both issues by i) computing the inner product of the quartic of the IDF scores so to create even more contrast between common and rare words and ii) calculating the similarity between the query and the response. This is formalized as:
\begin{equation}
\mathcal{S}^{cr} =  q^4 \cdot r^4_i
\end{equation}

It can also be useful to consider a wider context $c_i$ as well as more than one previous user utterances when choosing a response.\\ $\mathcal{S}^{2cq}$ does that by considering as context $r_{i-2}$ as well as the agent response preceding $q_c$. For this subscore, normal IDF scores are used. This translates into the following equation:
\begin{equation}
\mathcal{S}^{2cq} = q_c \cdot c_{i}, \text{where } c_i = r_{i-2}
\end{equation}
In the case of ties or near ties, we select the argmax of $\lVert r_i \rVert_2$ to try to select a response that is longer and with more rare words.

Finally, the matching score keeps track of the most recent topic or subtopic. When a response is pulled from a particular sub-dataset from our AMT data, the sub-dataset topic is assumed to be the current topic. All responses from the sub-dataset covering the current topic are given a small bonus to their score to improve topic coherence. The matching score returns the highest scoring response as well as the score of that response, which we refer to as the confidence score.

\subsection{Likes and dislikes}
This component is triggered only if the rule-based agent fails to return and the matching score confidence drops below a predefined threshold. It contains a set of very specific templates that depend on the number of recent views of the entity's Wikipedia page. These templates are limited to `safe' domains, e.g. actors, singers and bands, to avoid sensitive entities. It is particularly useful for unseen entities. An example response by this component is shown below in (\ref{egs:edinaslikes}).
\eenumsentence{\label{egs:edinaslikes}
\item[] User: Do you like Kaizers Orchestra?
\item[] Edina: Kaizers Orchestra was a Norwegian alternative rock band formed on 1 January 2000. What do you think about them?}
                  
\subsection{Proactive component}
The proactive component is designed to drive the conversation towards a state that other components are better able to handle. It does so by sampling from a set of questions to ask the user. These questions are designed to either:
\begin{itemize}
\item probe the user to mention a specific entity that the Matching Score component will likely be able to match;
\item ask a yes or no question to gauge the user's interest in a particular topic, followed by a question probing the user to continue the conversation within that topic (cf. Example~\ref{egs:proactive}).
\end{itemize}
The proactive component is called when the matching score fails to return a response due to low confidence. It is in general beneficial if we can probe the user to mention specific entities related to the topics in our data, as our Matching Score component can often give interesting opinions about specific entities.  The proactive component is limited to only returning a response once very 5 turns to avoid excessive use. An example interaction is shown below.
\eenumsentence{\label{egs:proactive}
\addtolength{\itemsep}{-5pt}
\item[]Edina: Did you enjoy the last Star Wars movie?
\item[]Customer: Yes 
\item[]Edina: I was hoping for you to say yes! What did you think of it?}

\subsection{Generative Neural Network component}
\label{sec:nn}
The generative neural network is a recurrent neural network (RNN) language model that can predict a probability over the next word given a history of words. For this component, we used a multiplicative LSTM \citep{krause2017} which combines features of a multiplicative RNN \citep{sutskever2011} and an LSTM ~\citep{hochreiter1997long} to achieve a stronger language modeling performance. We pretrained our network on the open subtitles dataset ~\citep{Lison2016} using vocabulary from our self-dialogues. We then fine-tuned this model on our data sources. We considered several stopping points to try to obtain the model that generates the best samples. In general, some degree of overfitting (continuing to train even after validation error is getting worse) was beneficial to sample quality. At runtime, we simply sampled this model word by word, conditioned on the user's last response, and the agent's response before that. We used a temperature of 0.7 for sampling, to draw samples from higher density regions under the model. 

\section{System Evaluation}
\label{sec:eval}
\subsection{Internal evaluation of matching score}
We evaluated our matching score component and its associated confidence scores using an internal human evaluation metric. A single trial in our evaluation process involved sampling a triple $(r_{i-2},r_{i-1},r_i)$ from our data-bank of self-dialogues. We then use this conversational context as the user utterance for our matching score ($q_c=r_{i-2}$ and $q=r_{i-1}$). We then find the top 4 responses under our scoring function $\mathcal{S}(q,r_j,c_j)$, for $i \neq j$ , forcing the matching score to chose a response other than the true response $r_i$ which was given by the AMT worker. We randomly sampled response $r$ to either be one of these top-4 matching score responses, or the response given by the AMT worker ($r=r_i$). The response $r$ was then rated in the context of $q_c$ and $q$ according to the evaluation metric given below. The rater was always blind to whether $r$ was a real AMT worker response.

\textbf{Evaluation scale} \\
5 - human-like response that is interesting or engaging \\
4 - human-like response that is not as interesting or engaging, or response that doesn't quite fit conversationally but is interesting or engaging\\
3 - probably not a human-like response, but still somehow addresses part of the conversation context\\
2 - something is clearly off about response, but still could contribute to the conversation in some way\\
1 - response is incoherent.

We collected 181 trials of human evaluations of top-4 matching score responses and 33 trials of human evaluations of real responses from AMT workers. We found that the confidence scores of our top-4 matching score responses were positively correlated with human evaluation scores ($r=0.259$, $p<0.0005$). Statistics on human evaluation of matching score at different confidence thresholds and real human responses are given in Table ~\ref{tab:humaneval}.

\begin{table}[tb]
\begin{center} 
\begin{tabular}{ l l } \toprule Response type & Mean human evaluation score \\ \midrule  Human AMT worker &  $4.09 \pm 0.15$ \\  Matching score top 4, confidence > 0.7 & $3.60 \pm 0.07$ \\  Matching score top-4, confidence > 0.5  & $3.50 \pm 0.08$ \\  Matching score top-4 & $3.08 \pm 0.09$  \\  \bottomrule 
\end{tabular} 
\end{center}
\caption{Average human evaluation of matching score and real human responses, plus or minus standard error. Confidence threshold of 0.5 corresponded to confidence threshold for matching score component to response. Confidence threshold of 0.7 corresponded to confidence threshold for matching score component to move ahead in the queue of the likes and dislikes component.}
\label{tab:humaneval}
\end{table}

\subsection{Qualitative evaluation on online customer conversations}

Following, we show an excerpt from an online conversation, where the agent's responses are annotated with the components that generated them:
\begin{quote}
Bot: Hi! This is an Alexa Prize socialbot. I feel very rock and roll today. Shall we talk about music? [StartMessage] \\
User: sure what do you want to talk about \\
Bot: Lets talk about music [matching score] \\
User: okay which music in particular \\
Bot: Who is your favorite singer? [proactive] \\
User: David bowie \\
Bot: He passed away recently, very sad. [matching score] \\
User: I agree do you like David bowie \\
Bot: I'm a major David Bowie fan! You too? [matching score] \\
User: sure what's your favorite song \\
Bot: Hit me baby one more time by Britney Spears. Hashtag classic. [rule based] \\
User: what's your favorite David bowie song \\
Bot: That's a tough question. I would have to say Hunky Dory or Station to Station. [matching score] \\
User: what's your favorite artist \\
Bot: I adore the Guns and Roses. Such a great music. [rule based] \\
User: what is your favorite song is hit me baby 1 more time by Britney spears \\
Bot: That's an odd but respectable classic pick. [matching score] \\
User: what's your favorite guns and roses song \\
Bot: I have to go with November Rain. [matching score] \\
User: why do you like guns and roses \\
Bot: I love Guns and Roses. [matching score]
\end{quote}

\textit{Edina} demonstrates many of the advantages of having a hybrid method. The matching score was able to return a \textbf{current} output recognizing the fact that David Bowie has passed away recently. The outputs are \textbf{relevant} to the context; for instance the matching score was able to name actual songs by David Bowie and Guns and Roses. The rule based component is instead evoked correctly when the customer inquiries the preferences of our bot. Finally it is worth noticing that when the confidence threshold is low, the proactive component probes the user for an entity related to the current topic so not to disrupt the flow of the conversation.

\section{Discussion}
\label{discussion}

\textit{Edina} demonstrated the ability to engage in a wide range of subtopics during conversations with Alexa customers. This was achieved by drawing responses from a conversation dataset created with the novel technique of \textit{self-dialogues}. This technique plays a significant role in enabling our agent to generate relevant, topical output closely resembling human conversations. Our internal system evaluation further demonstrated the value of using a thresholded matching score to select a response. During our human evaluation, the responses that have a high matching confidence score received a significantly higher rating compared to the other responses (see Table~\ref{tab:humaneval}), showing that the matching score can self-censor to greatly reduce the risk of a poor response.

We believe the strength of our system is incorporating the advantages of both data-driven and rule-based approaches while avoiding their shortcomings. We were able to cover a broad range of topic and subtopics without the need of tailored scripts or rules, while at the same time using data that does not contain a large amount of noise and is easy to collect. Our self-conversation data collection technique and our unique approach to integrating rule-based, retrieval, and machine learning based methods should prove useful for future open domain conversational agents.

\bibliographystyle{plainnat}
\bibliography{sample}

\end{document}